\documentclass[10pt,twocolumn,letterpaper]{article}

\usepackage[pagenumbers]{cvpr} 
\usepackage[dvipsnames]{xcolor}
\usepackage{graphicx}
\usepackage[acronym,toc]{glossaries}
\usepackage[flushleft]{threeparttable}

\definecolor{cvprblue}{rgb}{0.21,0.49,0.74}
\usepackage[pagebackref,breaklinks,colorlinks,citecolor=cvprblue]{hyperref}

\def\paperID{*****} 
\def\confName{CVPR}
\def\confYear{2024}

\title{SUN Team's Contribution to ABAW 2024 Competition: Audio-visual Valence-Arousal Estimation and Expression Recognition}

\author{Denis Dresvyanskiy$^*$\\
Ulm University, Germany\\
ITMO University, Russia\\
{\tt\small denis.dresvyanskiy@uni-ulm.de}
\and
Maxim Markitantov$^*$\\
St. Petersburg Federal Research Center\\ of the Russian Academy of Sciences (SPC RAS), Russia\\
{\tt\small markitantov.m@iias.spb.su}
\and
Jiawei Yu\\
Information and Computing Sciences Dept.\\
Utrecht University, The Netherlands\\
{\tt\small j.yu@uu.nl}
\and
Peitong Li\\
Information and Computing Sciences Dept.\\
Utrecht University, The Netherlands\\
{\tt\small p.li1@uu.nl}
\and
Heysem Kaya\\
Information and Computing Sciences Dept.\\
Utrecht University, The Netherlands\\
{\tt\small h.kaya@uu.nl}
\and
Alexey Karpov\\
St. Petersburg Federal Research Center\\ of the Russian Academy of Sciences (SPC RAS), Russia\\
{\tt\small karpov@iias.spb.su}
}

\begin{document}
\maketitle
\def\thefootnote{*}\footnotetext{These authors contributed equally to this work}

\newacronym{CNN}{CNN}{Convolutional Neural Networks}
\newacronym{PDEM}{PDEM}{Public Dimensional Emotion Model}
\newacronym{DNN}{DNN}{Deep Neural Networks}
\newacronym{ABAW24}{ABAW'24}{Affective Behavior Analysis in-the-Wild 2024}
\newacronym{FER}{FER}{Facial Expression Recognition}
\newacronym{ViT}{ViT}{Visual Transformer}
\newacronym{RECOLA}{RECOLA}{Remote Collaborative and Affective Interactions}
\newacronym{FPS}{FPS}{Frames Per Second}
\newacronym{SER}{SER}{Speech Emotion Recognition}
\newacronym{GRU}{GRU}{Gated Recurrent Unit}
\newacronym{FCL}{FCL}{Fully Connected Layer}
\newacronym{CCC}{CCC}{Concordance Correlation Coefficient}
\newacronym{E2E}{E2E}{End-to-End}

\begin{abstract}
As emotions play a central role in human communication, automatic emotion recognition has attracted increasing attention in the last two decades. While multimodal systems enjoy high performances on lab-controlled data, they are still far from providing ecological validity on non-lab-controlled, namely `in-the-wild' data. This work investigates audiovisual deep learning approaches for emotion recognition in-the-wild problem. We particularly explore the effectiveness of architectures based on fine-tuned \acrfull{CNN} and \acrfull{PDEM}, for video and audio modality, respectively. We compare alternative temporal modeling and fusion strategies using the embeddings from these multi-stage trained modality-specific \acrfull{DNN}. We report results on the AffWild2 dataset under \acrfull{ABAW24} challenge protocol. 
\end{abstract}

\section{Introduction}
\label{sec:intro}
This paper presents out contribution to the 2024 edition~\cite{kollias20246th} of the Affective Behavior Analysis in-the-wild (ABAW) challenge series~\cite{kollias2023abaw2, kollias2023multi,  kollias2023abaw, kollias2022abaw_2, kollias2021analysing, kollias2021affect, kollias2021distribution, kollias2020analysing, kollias2019expression, kollias2019deep, kollias2019face, zafeiriou2017aff}. The challenges in the field of affective computing have boosted the development of state-of-the-art methods, while also ensuring the reproducibility and comparability of the developed methods under a common experimental protocol. 

This year, sub-challenges of ABAW include 8-class categorical emotion recognition (Expression Challenge- EXPR), featuring Ekman's six basic emotions (anger, disgust, fear, happiness, sadness, surprise), plus \textit{neutral} and \textit{others} classes as well as emotion primitives (arousal and valence) regression challenge (VA). We participated in these two sub-challenges. For the challenge data, its annotation process and further details, we refer the reader to~\cite{kollias20246th} and the former editions of ABAW. We continue with the details of the proposed approach in the next section.

\section{Methodology}
\label{sec:methodology}

\subsection{Acoustic Emotion Recognition System}
\label{subsec:acoustic_method}
We proposed three slightly different models. The backbone of all models is based on the \acrshort{PDEM}. The \gls{PDEM}, designed for predicting arousal, valence, and dominance, is the first publicly available transformer-based dimensional \gls{SER} model \cite{wagner2023dawn}. It is fine-tuned on the pre-trained wav2vec2-large-robust model, which is one of the variants of Wav2Vec 2.0~\cite{baevski2020wav2vec}.

On top of each model, we stack two \gls{GRU} layers with 256 neurons (AudioModelV1) or two transformer layers with self-attention mechanisms, each with 32 and 16 heads (AudioModelV1 and AudioModelV2). After the last transformer layer, we aggregate the information along the time axis using \gls{CNN} and apply two subsequent \glspl{FCL} for feature compression and for the classification or regression layer, depending on the challenge. We fine-tune all the layers from the top to the last two (AudioModelV1 and AudioModelV2) or four (AudioModelV3) encoding layers of the backbone model.

\subsection{Visual Emotion Recognition System}

For the visual emotion recognition system, we have trained the final model in several stages. First of all, we have pre-trained the modified static models (EfficientNet-B1, -B4, and ViT-B-16) on the pre-training datasets introduced in~\ref{subsubsection:pre-training}. Next, to further enhance the efficacy and robustness of the static \gls{FER} models, they have been fine-tuned on the AffWild2 dataset. Finally, the fine-tuned static models have been frozen and used as accurate feature extractors that provide valuable affective features for consecutive temporal aggregation using the dynamic \gls{FER} model. In the next subsections, we describe the parts of the pipeline of the final visual dynamic \gls{FER} system.

\subsubsection{Static Models}
To construct an accurate emotion recognition model, especially for the visual data, a robust and efficient feature extractor is needed. Those models are called static as they are trained on and provide emotion predictions as well as informative facial features per-frame, not paying attention to the temporal context.
In the context of \gls{FER}, such state-of-the-art models are based either on \acrshort{CNN} or recently introduced \gls{ViT} neural network architectures. In this work, we experimented with both approaches, as various models can demonstrate different performances given in-the-wild nature of the data. Specifically, we have employed the EfficientNet~\cite{EfficientNet} (B1 and B4 versions, comprising 7.8M and 19.3M parameters, respectively) and Visual Transformer-B-16~\cite{dosovitskiy2021an} architectures that are pre-trained on ImageNet~\cite{DengImagenet09,russakovsky2015imagenet}. 
\begin{figure}
	\centering
	\includegraphics[width=0.99\columnwidth]{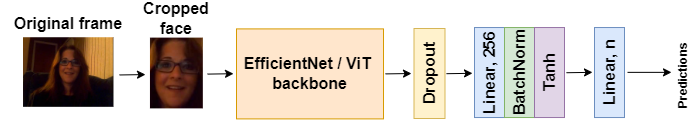}
	\caption{The neural network architecture of modified frame-level FER models. \textit{n} -- number of neurons, 8 for classification task and 2 for regression task. After generation of predictions, a \textit{softmax} or \textit{tanh} activation function is applied depending on the task type.}
	\label{fig:efficientnet}
\end{figure}

However, before fine-tuning those models on various emotion recognition datasets (including AffWild2), we have slightly modified them as depicted in Figure~\ref{fig:efficientnet}. Thus, we removed the last layer responsible for the classification and stacked on top of it (1) the dropout followed by deep embeddings layer (feed-forward) with 256 neurons, batch normalization, and \textit{Tanh} activation function, (2) the new classification or regression layer with the corresponding number of neurons (8 for the classification task, 2 for arousal-valence regression). For classification task, a softmax activation function is used, while for regression (Valence and Arousal) we used the \textit{Tanh} as it transforms the output values into the [-1, 1] range. 

\subsubsection{Dynamic Models}

\begin{figure}
	\centering
	\includegraphics[width=0.99\columnwidth]{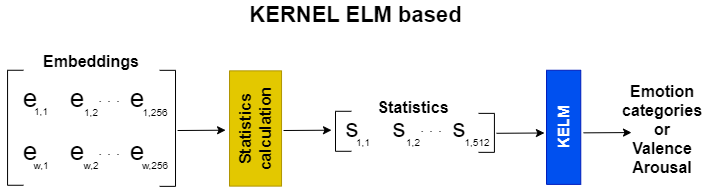}
	\caption{The architecture of Kernel ELM based dynamic emotion recognition model.}
	\label{fig:dynamic_architectures}
\end{figure}

\begin{figure*}
	\centering
	\includegraphics[width=0.99\textwidth]{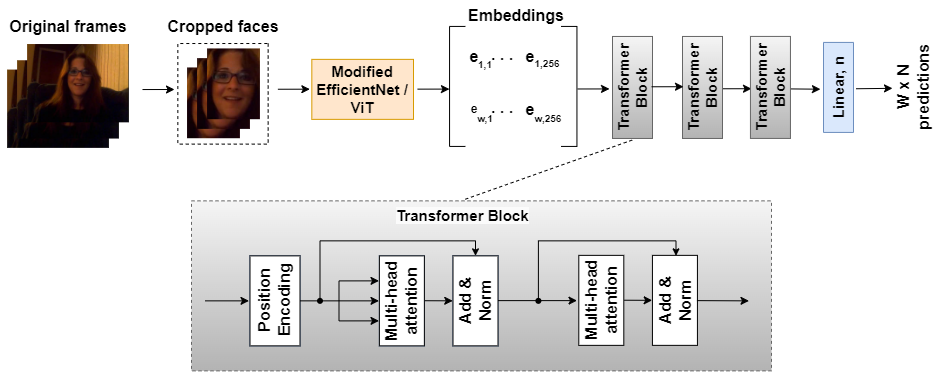}
	\caption{Pipeline and architecture of the transformer-based sequence to one video emotion recognition model. $W$ -- the temporal window size (in the number of frames), $N$ -- the number of neurons in the decision-making head (either 8 for classification or 2 for regression task).}
	\label{fig:dynamic_transformer}
\end{figure*}

It is well-known that emotions are temporal phenomena. They last for a certain period and are reflected in the facial, vocal, and body dynamics. To exploit this aspect, we developed dynamic \gls{FER} models that take into account the temporal context during the decision-making process. An overview of these model architectures is given in Figures~\ref{fig:dynamic_architectures} and ~\ref{fig:dynamic_transformer}.

In the emotion recognition literature, there are many different approaches for temporal information aggregation, including functionals-aggregation (calculation of statistics over period of time), recurrent neural networks~\cite{DresvyanskiyMTI22}, and recently introduced Transformers-based architectures~\cite{lian2021ctnet,chaudhari2022vitfer,WagnerPDEM23}. Although recurrent neural networks have been most popular in \gls{FER} domain so far, the Transformers-based architectures are taking the lead in the last years. 

To leverage the most effective architectures and build a robust \gls{FER} model, we employed the Transformer-based temporal aggregation method as well. The implemented \gls{FER} dynamic model is schematically depicted in Figure~\ref{fig:dynamic_transformer}. 
Thus, the dynamic model consists of a static feature extractor and the temporal part that consists of three consecutive Transformer-encoder-based layers inspired by~\cite{VaswaniNIPS17_Transformer}. Lastly, the classification or regression head completes the decision-making process. For the regression case, a \textit{Tanh} activation function is used.

We should note that the feature extractor is firstly pre-trained on several \gls{FER} datasets, then fine-tuned on the AffWild2 dataset frames, and is finally frozen so that it does not change its weights during the training of the dynamic model. As we fixed the number of static embeddings by modifying respective static models, the feature extractor always outputs 256 features per frame. For the Transformer-based layers, we set the number of heads equal to 8 and dropout equals 0.1. Additionally, the positional encoding employed in~\cite{VaswaniNIPS17_Transformer} is applied to visual embeddings. 

For the comparison and as an alternative, we developed a simpler temporal aggregation method: the statistical-based model that calculates functionals over a fixed period. Here, based on previous work on an earlier version of ABAW challenge~\cite{DresvyanskiyMTI22}, we fix the analysis window to 2 seconds. We apply mean, minimum, and maximum functional statistics to non-overlapping 2-second windows and apply Kernel Extreme Learning Machine (KELM)~\cite{huang2012extreme}. KELM aims to solve a regularized least squares regression problem between a kernel (instance similarity) matrix \(\textbf{K}\) and a target vector (or matrix) \(\textbf{T}\) from the training dataset, and hence is very fast to train given the kernel. The set of weights ($\beta$) in KELM is calculated via:
\begin{equation}
	\beta = (\textbf{I}/C + \textbf{K})^{-1}\textbf{T},
	\label{eq:3}
\end{equation}
where \(\textbf{I}\) is the identity matrix, and \(C\) is the regularization coefficient optimized via cross-validation on the challenge development set. The prediction for a test instance \(x\) is obtained via \(\hat{y}={K}(\textbf{D},x)\beta\), where \({K}(,)\) and \(\textbf{D}\) denote the kernel function and the training dataset, respectively. KELM is used both for regression and classification in this study. For the classification (Expr) task, considering the challenge performance measure and the class imbalance, we use the Weighted KELM~\cite{Zong13_weightedELM} that weights each instance inversely proportional to the training set instance count of that class. This extension is not only more memory-efficient compared to upsampling but also found to yield more accurate results in challenging audio~\cite{kaya17_interspeech} and video-based~\cite{baki2022multimodal} recognition tasks.

\subsection{Fusion Schemes}
Fusion, particularly to leverage complementary multi-modal information is an important stage in video-based emotion recognition systems. In our experiments, we experimented with decision (late) and model-based fusion strategies. In the latter, the features from audio and video models are concatenated and fed to attention mechanisms (self-attention and cross-attention). 

For late fusion, we experimented with two schemes. First, we used Dirichlet-based Random Weighted Fusion (DWF), where fusion matrices containing weights per model-class combination are randomly sampled from the Dirichlet distribution. A large pool of such matrices is generated and the matrix that gives the best performance in terms of the task-wise challenge measure is selected for the test set submission. This approach is shown to generalize well to in-the-wild emotion recognition in former studies~\cite{KAYA_Imavis17, DresvyanskiyMTI22}.

The second decision fusion approach is based on Random Forests (RF)~\cite{breiman1996bagging}, where the concatenated probability vectors from the base models are stacked to RF. To avoid over-fitting, out-of-bag predictions are probed to optimize the number of trees.

\section{Experimental Setup}
\label{sec:experimental_setup}
\subsection{Experimental Data}
For all experiments presented in this work, the AffWild2 dataset and corresponding labels from the 6th ABAW challenge~\cite{kollias20246th} have been used. 

The AffWild2 dataset is an audiovisual in-the-wild corpus, that serves as a comprehensive benchmark for multiple affective behavior analysis tasks and is utilized in ABAW Competition series. Comprising 594 videos with approximately 3 million frames from 584 subjects, it is annotated in terms of Valence and Arousal emotional continuous labels ranging from -1 to 1. Additionally, a subset of 548 videos (approximately 2.7 million frames) is annotated for expression recognition across 8 classes, including 6 Ekman's basic expressions, the Neutral state, and \textit{Other}. 

The dataset is subject-independently partitioned into training, validation, and test sets, ensuring no subject overlap across partitions. Thus, the AffWild2 dataset poses several emotion recognition challenges to be solved, advancing affective computing research in unconstrained settings.

\begin{table*}[!ht]
	\begin{center}
		\centering
		\caption{The summary of pre-training corpora used in this work. Note that in the ``Data volume'' column, only the volume of data utilized in this work is presented.}
		\label{table:chapter_3_datasets}.
		\begin{tabular}{l|cccc}
			\hline
			\textbf{Dataset} & \textbf{Modality} & \textbf{Data volume} & \textbf{Annotations} & \textbf{Conditions}\\
			\hline
			RECOLA~\cite{ringeval2013introducing} & Audio, Visual & 3:50 hours & A, V & In-the-wild \\
			SEWA~\cite{kossaifi2019sewa} & Audio, Visual & 9:10 hours &  A, V & In-the-wild \\
			SEMAINE~\cite{mckeown2011semaine} & Audio, Visual & 6:30 hours &  A, V & Lab. \\
			AFEW-VA~\cite{kossaifi2017afew_va} & Visual & $\approx$30,000 frames &  A, V & In-the-wild \\
			AffectNet~\cite{mollahosseini2017affectnet} & Visual & 420,299 images & C, A, V & In-the-wild \\
			SAVEE~\cite{jackson2014surrey} & Audio, Visual & $\approx$24 minutes & C & Lab. \\
			EMOTIC~\cite{Kosti_2017_CVPR_Workshops} & Visual & 23,571 images & C, A, V & In-the-wild \\
			ExpW~\cite{zhang2018facial} & Visual & 91,793 images & C & In-the-wild \\
			FER+~\cite{goodfellow2013challenges} & Visual & 35,887 images & C & Lab. \\
			RAF-DB~\cite{Li_2017_CVPR} & Visual & 29,672 images & C & In-the-wild \\
			\hline
		\end{tabular}
		
		\begin{tablenotes}
			\centering
			\item A -- Arousal, V -- Valence, C -- Categories
			\item Lab. - Laboratory conditions
		\end{tablenotes}
	\end{center}
\end{table*}

\subsubsection{Pre-training Data}
\label{subsubsection:pre-training}
To pre-train our static \gls{FER} models, we used a range of publicly available datasets, namely, \gls{RECOLA}~\cite{ringeval2013introducing}, ~\cite{kossaifi2019sewa}, SEMAINE~\cite{mckeown2011semaine}, AFEW-VA~\cite{kossaifi2017afew_va} (which is based on AFEW~\cite{dhall2012collecting}), AffectNet~\cite{mollahosseini2017affectnet}, SAVEE~\cite{jackson2014surrey}, EMOTIC~\cite{Kosti_2017_CVPR_Workshops}, ExpW~\cite{zhang2018facial}, FER+~\cite{goodfellow2013challenges}, RAF-DB~\cite{Li_2017_CVPR}.
To pre-train those models, Ekman's six basic emotions were selected from the aforementioned datasets along with Valence and Arousal values. We summarized the information about all used corpora in Table~\ref{table:chapter_3_datasets}. 

After the pre-training phase, the static \gls{FER} models have been fine-tuned on the AffWild2 dataset.

\subsection{Data Preprocessing}

\subsubsection{Audio}
Before training an audio model, in addition to extracting audio signals from multimedia files, we perform voice activity detection. Due to the specific nature of the acoustic data provided by the ABAW24 challenge organizers, audio data may include background noise and multiple speakers, making it difficult to identify the target speaker, methods based only on audio analysis are not suitable. Therefore, we rely on video modality by analyzing the video data frame by frame. For this purpose, facial landmarks are extracted using the MediaPipe framework~\cite{lugaresi2019mediapipe}, then mouth landmarks are detected, and the corresponding region of interest is extracted. It is used to determine whether the target speaker's mouth is open or closed. Simultaneously, we filter the acoustic data from noise using Spleeter by deezer\footnote{https://github.com/deezer/spleeter}.

After that, we downsample all videos with varying frame rates to 5 \gls{FPS}, which helps to fix the window length. Then, 4-second windows with a step of two seconds are formed on the filtered detected voice segments. In case of the EXPR challenge, to obtain the target label at each second, we compute the most frequent frame-wise label. In case of the VA challenge, we utilize all downsampled values. Therefore, each window has four labels and 20 valence/arousal values for the EXPR and VA challenges, respectively.

To enhance the generalizability of the audio models, we employ several augmentation techniques, including polarity inversion, the addition of white noise or variation in audio volume, and Label Smoothing~\cite{lukasik2020does}. These techniques help to reduce the confidence level of the models in their emotion predictions.

\subsubsection{Video}
\begin{figure}
	\centering
	\includegraphics[width=0.99\columnwidth]{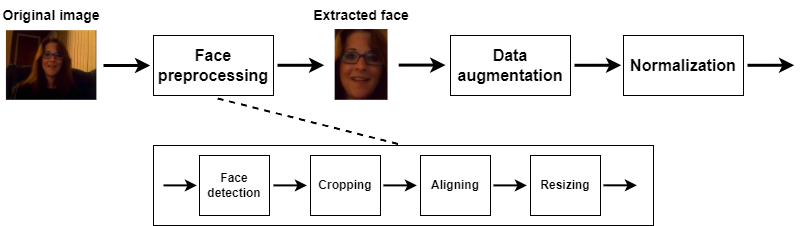}
	\caption{Pipeline of the image preprocessing for the static \gls{FER} models.}
	\label{fig:staticpreprocess}
\end{figure}

\begin{figure*}[!ht]
	\centering
	\includegraphics[width=0.7\textwidth]{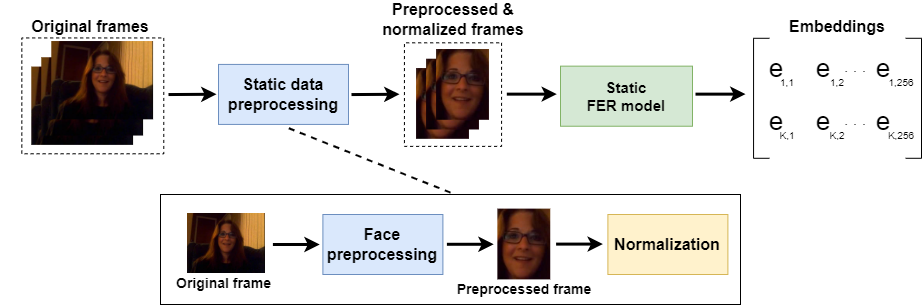}
	\caption{Pipeline of preprocessing of the video data for dynamic emotion recognition modeling.}
	\label{fig:dynamicpreprocess}
\end{figure*}
Depending on the model type (static or dynamic), several preprocessing steps have been applied. 
For the static \gls{FER} model, we have depicted the preprocessing pipeline in Figure~\ref{fig:dynamicpreprocess}. We first detect faces and crop them, adding 15 pixels to all bounding box boundaries to include the chin and other human facial features. we have utilized the RetinaFace model~\cite{deng1905retinaface,9157330} based on the MobileNet \cite{MobileNet} architecture, namely the \textit{MobileNet-0.25} version. We have chosen this model because it is one of the most effective face recognition models known nowadays, yet very computationally efficient, since it has only around 1.7 million parameters. 
The next step in the static data preprocessing pipeline is to resize the image and normalize the pixel values. In this work, we employed the needed image resolutions and normalization values provided by the authors of corresponding models (EfficientNet-B1~\cite{EfficientNet} and ViT-B-16~\cite{dosovitskiy2021an}). 
Finally, to improve the performance and robustness of the deep learning model, various data augmentation techniques are applied. During the training of the static emotion recognition models, we have used the following image augmentation techniques: random image padding, grayscaling, random changing of brightness, contrast, saturation, and hue of the image, noise addition (gaussian noise), random rotation, random cropping, image posterization, changing of sharpness, equalization, and flipping. All augmentation techniques were applied to the image with a probability equal to 0.05. Thus, approximately 46\% of images have been augmented every training epoch.

In the context of the dynamic \gls{FER} model, the data preprocessing pipeline is depicted in Figure~\ref{fig:dynamicpreprocess}. It closely mirrors the static methodology except for one important step: we use additional normalization of embeddings to avoid the gradient explosion that can arise in the early stages of training. Two different normalization methods have been used: MinMax scaling and Per-video-MinMax scaling. The difference in methods is that MinMax scaling computes the corresponding min and max values across the whole training set (and then applied to every instance), while the Per-video-MinMax scaling computes min and max values for every video separately (and applies those values only within the corresponding video).

It is well-known that Transformer-based models can operate with sequences of arbitrary length. However, different video frame rates (FPS) of videos and varying lengths of the video sequences can significantly harm the training process. Therefore, to stabilize it and ensure convergence, we downsampled all videos with varying frame rates to 5 FPS and fixed the window length. It is experimentally shown that the size of a temporal window can significantly influence the efficacy of the \gls{FER} model~\cite{DresvyanskiyMTI22}. That is why we have done experiments with different temporal context lengths, namely: 1, 2, 3, 4, 6, and 8 seconds. Finally, the model with the highest competition measure value is used for test set submissions.

\subsubsection{Post-processing}
After getting class probabilities or Valence and Arousal values for every frame, several post-processing steps are applied. Since dynamic models are trained using reduced FPS, we first interpolated the models' predictions to align them with the ground truth labels on the validation set and the actual video FPS on the test set. We used linear interpolation, filling in missing values between two consecutive predictions. After interpolation, we applied the smoothing using Hamming window~\cite{tukey1958measurement}. The size of the window has been chosen to be 0.5 seconds.

\section{Experimental Results}
\label{sec:experimental_results}
The challenge measures for expression (EXPR) and dimensional emotion (VA) challenges are F1 and \gls{CCC}, respectively. \gls{CCC} is recently popularly used in regression tasks over the \textit{Pearson's Correlation} (PC), as it also considers the difference in means~\cite{lawrence1989concordance}:
\begin{equation}
	CCC = \frac{2\cdot \sigma_{t,p}}{\sigma_{t}^2+\sigma_{p}^2+(\mu_{t}-\mu_{p})^2},
\end{equation}
where $\mu_{t}$ and $\mu_{p}$ denote the averaged ground truth and predicted scores for all test clips, respectively; $\sigma_{t}$ and $\sigma_{p}$ denote the respective standard deviations; $\sigma_{t,p}$ is the covariance between $t$ and $p$.

\subsection{Audio-based Models}
For acoustic modality, we generated three base models via \gls{E2E} fine-tuning of the \gls{PDEM} model. All of the top approaches used data augmentation and Spleeter for background noise separation. Note that the results reported here are on 4-second windows excluding the unvoiced segments, rather than frame-wise over which the ground truth annotations are provided. The best development set performance for the EXPR and VA challenges are presented in Tables~\ref{tab:audio_expression} and~\ref{tab:aud_va}, respectively. On both tasks, the best development set performance is obtained with AudioModelV3. We therefore use (and fuse) these models' predictions for our test set submissions.

\begin{table}[ht]
	\caption{Best development set results per acoustic base-model on the EXPR challenge.}
	\label{tab:audio_expression}
	\centering
	\begin{tabular}{lrrr}
		\hline
		Model       & F1    & Recall & Precision   \\
		\hline
		AudioModelV3 & \textbf{0.347} & 0.350  & 0.376      \\
		AudioModelV2 & 0.335 & 0.342  & 0.343      \\
		AudioModelV1 & 0.319 & 0.312  & 0.341     \\
		\hline
	\end{tabular}
\end{table}

\begin{table}[ht]
	\caption{Best development set results per acoustic base-model on the VA challenge.}
	\centering
	\label{tab:aud_va}
	\begin{tabular}{lrrr}
		\hline
		Model     & Arousal & Valence & Average   \\
		\hline
		AudioModelV3 & 0.400   & 0.290   &  \textbf{0.345}     \\
		AudioModelV1 & 0.377   & 0.282   & 0.329    \\
		AudioModelV2 & 0.375   & 0.241   & 0.308   \\
		\hline
	\end{tabular}
\end{table}

\subsection{Video-based Models}
We experimented with EfficientNet-B1 and ViT models for extracting features for subsequent modeling. Using the functionals-based approach, where the embeddings are summarized over 2-second non-overlapping windows, we obtained decent results on both EXPR (see Table~\ref{tab:vid_func_expr}) and VA. Based on the results of the EXPR challenge, we optimized the VA model using the combination of mean, min, and max functionals. The best development set \gls{CCC} (0.489 average over two dimensions) was obtained using EfficientNetB1 model, with corresponding CCC performances of 0.398 and 0.581 for valence and arousal, respectively. Note that the results reported for the functionals-based approach are on 2-second windows, rather than frame-wise.

\begin{table}[ht]
	\caption{Best development set results using Functionals-based video summarization approach. Embeddings are extracted from fine-tuned ViT model.}
	\label{tab:vid_func_expr}
	\centering
	\begin{tabular}{lrrrr}
		\hline
		Functionals            & {F1} & {Acc.} & {Precision} & {Recall} \\
		\hline
		mean, max, min & \textbf{0.354}                  & 0.478                        & 0.338                         & 0.416                     \\
		mean, max      & 0.344                  & 0.462                        & 0.328                         & 0.412                     \\
		mean, min      & 0.352                  & 0.469                        & 0.336                         & 0.419 \\
		\hline
	\end{tabular}
\end{table}

Extensive experiments with end-to-end models showed the sensitivity of these models to hyperparameters. These models' predictions are post-processed to match the ground truth label frequency. Our best performance (0.387 F1 score) on the EXPR challenge was obtained with ViT as an embedding extractor, followed by three Transformer layers for temporal modeling. The best E2E video model on the VA task was, on the other hand, EfficientNet-B1, reaching an average CCC performance of 0.574, with corresponding arousal and valence CCC scores of 0.626 and 0.523, respectively. These along with the best functional-based system's predictions are later used for test set predictions.

\subsection{Multimodal Models and Test Submissions}
\begin{table*}[ht]
	\caption{Development set CCC performances of the submitted systems for the VA challenge. DWF: Dirichlet-based Random Weighted Fusion, RF: Random Forest-based fusion.}
	\label{tab:sub_va}
	\begin{tabular}{rllrrr}
		\hline
		{Sys \#} & Modality     & Method                                                      & {Average} & {Valence} & {Arousal} \\ \hline
		1                          & Visual       & Face-based E2E model (EfficientNet-B1 + Transformer layers) & 0.574                       & 0.523                          & 0.626                          \\
		2                          & Audio-visual & Sys1 + Best Audio (DWF)                                      & 0.591                       & 0.532                          & 0.649                          \\
		3                          & Audio-visual & Sys1 + Best audio + Best Functional  (DWF)                    & 0.584                       & 0.528                          & 0.641                          \\
		4                          & Audio-visual & Sys1 + Best audio + Best Functional (RF)              & 0.763                       & 0.738                          & 0.787                      \\ 
		\hline
	\end{tabular}
\end{table*}

\begin{table*}[ht]
	\centering
	\caption{Development set performances of the submitted systems for the EXPR challenge. DWF: Dirichlet-based Random Weighted Fusion, RF: Random Forest-based fusion.}
	\label{tab:sub_expr}
	\begin{tabular}{rllr}
		\hline
		{Sys \#} & Modality     & Method                   & {F1} \\
		\hline
		1                          & Visual       & Face-based E2E model (ViT + Transformer layers) & 0.397                               \\
		2                          & Audio-visual & Sys1 + Best Audio (DWF)                         & 0.458                              \\
		3                          & Audio-visual & Sys1 + Best Audio + Best Functional (DWF)             & 0.457                                 \\
		4                          & Audio-visual & Sys1 + Best Audio + Best Functional (RF)              & 0.999           \\ \hline            
	\end{tabular}
\end{table*}
We selected the best-performing audio and visual models, extracted embeddings, and experimented with the attention mechanism. This approach did not outperform the video-only system on the development set. Therefore, we used decision-based fusion schemes described earlier for the fusion. For our test set probes, we used one unimodal (best face-based E2E system) and three multimodal systems, based on the development set performances. By the time of writing, the test set results are not available to the challenge participants. Table~\ref{tab:sub_va} reports the development set performances of the VA challenge, whereas Table~\ref{tab:sub_expr} reports the corresponding development set performances of the EXPR challenge submissions. From~\ref{tab:sub_expr}, we see that the RF-based fusion overfits, even though a small number of trees (10) is used for fusion. For the sake of completeness, we submit predictions using both RF and DWF late fusion schemes.

\section{Conclusion and Future Work}
\label{sec:conclusion_future_work}

The results of our research highlight the potential of deep learning models for audiovisual emotion recognition in unconstrained, "in-the-wild" settings. The face-based end-to-end dynamic models, leveraging salient embeddings extracted by Visual Transformer and EfficientNet-B1 model, achieved a competitive efficacy, outperforming the traditional functional-based approaches on the AffWild2 dataset. However, optimizing video models still remains computationally very demanding, posing additional challenges in deploying these solutions for "in-the-wild" scenarios. 

While our experiments on the development set suggested that combining audio and video modalities through fusion techniques could significantly enhance performance, we do not have an opportunity to draw definitive conclusions about the advantages of multi-modal fusion approaches for the 6th ABAW challenge yet, as the test set results have not yet been revealed.

Looking ahead, further research could explore other novel fusion strategies that take into account additional contextual information such as background sound, participants' postures and gestures, linguistics, and other valuable complementary sources of information.
Ultimately, progress in this field holds promise for enhancing human-computer interaction and enabling a natural unconstrained usage of computer systems and robots in human society.

{
    \small
    
}


\end{document}